\documentclass[journal,twoside]{IEEEtran}
\usepackage[caption=false,font=footnotesize]{subfig}
\usepackage{cite}
\usepackage{fancyhdr}
\usepackage{algorithm}
\usepackage{algpseudocode}
\usepackage{algorithmicx}
\usepackage{multirow}
\usepackage{tabularx}
\usepackage{graphicx}
\usepackage{enumitem}
\usepackage{epstopdf}
\usepackage{ctable}
\usepackage{amsmath}
\usepackage{amsthm}
\usepackage{lineno}

\restylefloat{table}
\usepackage{array}

\begin{document}
\title{A New Intelligence Based Approach for Computer-Aided Diagnosis of Dengue Fever}
\author{Vadrevu Sree Hari Rao,~\IEEEmembership{Senior Member,~IEEE,} and Mallenahalli Naresh Kumar
\thanks{Manuscript received May 13, 2011: revised August 24, 2011 and September 30, 2011; accepted October 7, 2011.}
\thanks{Vadrevu Sree Hari Rao is with the Department of Mathematics, Jawaharlal Nehru Technological University, Hyderabad, Andhra Pradesh, 500 085, India. Also, he is an advisor for International Centre for Interdisciplinary Research and Innovation, VNRVJIET Campus, Hyderabad, India. e-mail: vshrao@jntuh.ac.in}
\thanks{Mallenahalli Naresh Kumar is with the Software and Database Systems Group, National Remote Sensing Center (ISRO),  Hyderabad, Andhra Pradesh, 500 625, India. e-mail: nareshkumar\_m@nrsc.gov.in}
\thanks{$\copyright$ 20xx IEEE. Personal use of this material is permitted. Permission from IEEE must be obtained for all other uses, in any current or future media, including reprinting/republishing this material for advertising or promotional purposes, creating new collective works, for resale or redistribution to servers or lists, or reuse of any copyrighted component of this work in other works. DOI: 10.1109/TITB.2011.2171978}
}

\maketitle
\markboth{\scriptsize $\copyright$ 20xx IEEE. Personal use of this material is permitted. Published in IEEE.  DOI: 10.1109/TITB.2011.2171978} {\scriptsize \thepage}
\begin{abstract}
Identification of the influential clinical symptoms and laboratory features that help in the diagnosis of dengue fever in early phase of the illness would aid in designing effective public health management and virological surveillance strategies. Keeping this as our main objective we develop in this paper, a new computational intelligence based methodology that predicts the diagnosis in real time, minimizing the number of false positives and false negatives. Our methodology consists of three major components (i) a novel missing value imputation procedure that can be applied on any data set consisting of categorical (nominal) and/or numeric (real or integer) (ii) a wrapper based features selection method with genetic search for extracting a subset of most influential symptoms that can diagnose the illness and (iii) an alternating decision tree method that employs boosting for generating highly accurate decision rules. The predictive models developed using our methodology are found to be more accurate than the state-of-the-art methodologies used in the diagnosis of the dengue fever.
\end{abstract}
\section{Introduction}

\IEEEPARstart{D}{engue} fever (DF) is a mosquito-borne infectious disease caused by the viruses of the genus \emph{Togaviridae}, subgenus \emph{Flavirus}. The transmission of this disease is through the bites of vectors (aedes aegypti, aedes albopictus) carrying the viruses belonging to \emph{Flavi genus}~\cite{Gubler1998}. From its first appearance in the Philippines in $1953$, the disease has been identified as one of the most important arthropod-borne viral disease in humans~\cite{monath1994}. Dengue virus infection has been reported in more than $100$ countries, with $2.5$ billion people living in areas where dengue is endemic. The annual occurrence is estimated to be around $100$ million cases of DF and $250,000$ cases of dengue hemorrhagic fever (DHF).

The diagnosis of dengue fever presents great challenges as the symptoms overlap with other febrile illnesses. Accurate diagnosis is possible only after conducting definitive tests such as enzyme-linked immunosorbent assays (ELISA) and real-time polymerase-chain reaction (RT-PCR) which are based on nucleic and acid hybridization~\cite{DePaula2004}. A recent study~\cite{Chen2008} on the behavior of C-type lectin domain family $5$, member A (CLEC5A) gene may result in a strategy for reducing tissue damage which would help improve the odds of survival of the patients suffering from DHF and dengue shock syndrome (DSS). A multivariate model was developed in~\cite{Ibrahim2004} for predicting hemoglobin (Hb) using predictors such as reactance obtained from a single frequency bioelectrical impedance analysis, sex, nausea/vomiting sensation and weight. These strategies can be employed only after $2-12$ days from the onset of the illness and require state-of-the-art laboratory facilities.

The World Health Organization (WHO) has arrived at a classification scheme for identifying the infected individuals based on clinical symptoms and laboratory features. The development of predictive models for diagnosis of dengue fever based on these schemes is affected by missing or incomplete data records in the clinical databases~\cite{ColleenMNorris2000} which may arise due to any or all of the following reasons (i) value being lost (erased or deleted) (ii) not recorded (iii) incorrect measurements (iv) equipment errors and (v) an expert not attaching any importance to a particular clinical procedure. Usually data is not collected from an organized research point of view~\cite{Cios2002}. The presence of large number of clinical symptoms and laboratory features requires one to search large sub spaces for optimal feature subsets. These issues unless addressed appropriately would hinder the development of accurate and computationally effective diagnostic system.

In view of the above challenges, we present the following novel features of our work:
\begin{itemize}
\item to identify the missing values (MV) in the data set and impute them by using a newly developed novel imputation procedure;
\item to identify a set of clinical symptoms that would enable early detection of suspected dengue in children and adults, which reduces the risk of transmission of the dengue fever in the community;
\item to identify the laboratory features and clinical symptoms that would enable better diagnosis and understanding of the disease in suspected dengue individuals. This renders optimal utilization of the laboratory resources required for confirmed diagnosis;
\item to build a predictive model that has a capability of rendering effective diagnosis in realtime. Further we compare its performance with other state-of-the-art methods used in the diagnosis of dengue fever.
\end{itemize}
The present paper is organized as follows: A survey of the state-of-the-art techniques for the diagnosis of dengue fever is presented in Section~\ref{review}, while in Section~\ref{chap8:sec1} we describe our novel methodology for computer-aided clinical diagnosis of dengue. The performance evaluation of the methodologies is described in Section~\ref{evalmethods}. The description of the data sets and the experimental results are presented in Section~\ref{chap8:sec3}. We present a comparison of our new imputation methodology with other imputation methods in Section~\ref{perMV}. In Section \ref{chap8:sec4} we discuss the computational complexity of our new method. Comparison of our new methodology with other state-of-the-art methods forms the subject of Section \ref{chap8:sec5}. Conclusions and discussion are deferred to Section~\ref{chap8:sec6}.
\section{Survey of the state-of-the-art techniques for diagnosis of dengue fever}
\label{review}
Logistic regression method was employed to identify clinical symptoms and laboratory features in $381$ individuals, out of which $148$ were confirmed dengue~\cite{Chadwick2006}. The data records with missing values (MV) are ignored and are deleted from the data set. In~\cite{ramos2009}, the study was conducted on clinical records comprising of $341$ children and $597$ adults out of which $38$ and $107$ respectively were laboratory-confirmed positive dengue cases. In this study the data fields that are incomplete or inaccurate for all suspected dengue cases were replaced with the known values corresponding to the information in the medical charts. A C4.5 decision tree which has an in built mechanism of handling MV was employed in~\cite{Tanner2008} to develop a diagnostic algorithm to differentiate dengue from non-dengue illness on a data set comprising of $1200$ patients of which $173$ had DF, $171$ had DHF and $20$ had DSS.  A support vector machine (SVM) based methodology was employed in~\cite{gomes2010} to analyze the expression pattern of $12$ genes of $28$ dengue patients of which $13$ were DHF and $15$ were DF cases.  A set of seven influential genes were identified through selective removal of expression data of these twelve genes.

In the above studies the MV were either removed~\cite{Chadwick2006}, or filled with approximate values based on medical charts~\cite{ramos2009}. These approaches would lead to biased estimates and may either reduce or exaggerate the statistical power. Methods such as logistic regression, maximum likelihood and expectation maximization have been employed for imputation of MV, but they can be applied only on data sets that are either nominal or numeric. There are other imputation methods such as k-nearest neighbor imputation (KNNI)~\cite{BM2003}; k-means clustering imputation (KMI)~\cite{DSSL2004}; weighted k-nearest neighbor imputation (WKNNI)~\cite{TCSBHTBA2001} and fuzzy k-means clustering imputation (FKMI)~\cite{DSSL2004} that have been applied on other data sets but not on dengue fever data sets. However, the authors in~\cite{gomes2010,Chadwick2006,ramos2009} have employed methods such as odds ratio (OR) and selective inclusion or exclusion of attributes for obtaining features sub sets of data sets of dengue fever. But these methods do not yeild effective diagnosis as all interactions or correlations between the features and the diagnosis are not considered in these studies.
\section{A new methodology for computer-aided diagnosis of dengue fever}
\label{chap8:sec1}
Motivated by the above issues we propose a new methodology comprising of a novel non parametric missing value imputation method that can be applied on data sets consisting of attributes that are of the type categorical (nominal) and/or numeric (integer or real). The methodology proposed in~\cite{Freund1999} ignores missing values while generating the decision tree, which renders lower prediction accuracies. We have embedded the new imputation strategy (Section~\ref{chap9:RNI}) before generating the alternating decision tree which results in the improved performance of the classifier on data sets having missing values. Also, we develop an effective wrapper based features selection algorithm in order to identify the most influential features subset. The present methodology comprises in utilizing the new imputation embedded alternating decision tree and the wrapper based features subset selection algorithm. This methodology can predict the diagnosis of dengue in real time. In fact the machine knowledge acquired by utilizing this novel methodology will be useful to diagnose other individuals based on clinical symptoms and laboratory features where the clinical decision is unavailable. We designate this novel methodology as NM throughout this work.
\subsection{Data representation}
 A clinical data set can be represented as a set \emph{S} having row vectors  $(R_{1},R_{2},\hdots,R_{m})$ and column vectors $(C_{1},C_{2},\hdots,C_{n})$. Each record can be represented as an ordered n-tuple of clinical and laboratory attributes $(A_{i1}, A_{i2},\hdots, A_{i(n-1)}, A_{in})$ for each $\emph{i}=1,2,\hdots,m$ where the last attribute $(A_{in})$ for each \emph{i}, represents the physician's diagnosis to which the record $(A_{i1}, A_{i2},\hdots, A_{i(n-1)})$ belongs and without loss of generality we assume that there are no missing elements in this set. Each attribute of an element in \emph{S} that is $A_{ij}$ for $\emph{i}=1,2,\hdots,m$ and $\emph{j}=1,2,\hdots,n-1$ can either be a categorical (nominal) or numeric (real or integer) type. Clearly all the sets considered are finite sets.
\subsection{A new non-parametric imputation strategy}
\label{chap9:RNI}
The first step in any imputation algorithm is to compute the proximity measure in the feature space between the clinical  records to identify the nearest neighbors from where the values can be imputed. The most popular metric for quantifying the similarity between any two records is the Euclidean distance. Even though this metric is simpler to compute, it is sensitive to the scales of the features involved. Further it does not account for correlation between the features. Also, the categorical variables can only be quantified by counting measures which calls for the development of effective strategies for computing the similarity~\cite{Tadashi2009}. Considering these factors we first propose a new indexing measure $I_{C_{l}}(R_{i},R_{k})$ between two typical elements $R_{i}$, $R_{k}$ for $i,k=1,2,\ldots,m,$ $l=1,2,\ldots,n-1$ belonging to the column $C_{l}$ of \emph{S} which can be applied on any type of data, be it categorical (nominal) and/or numeric (real or integer). We consider the following cases:
\begin{description}
  \item[Case I:]~$A_{in}=A_{kn}$ \\
Let \emph{A} denote the collection of all members of \emph{S} that belong to the same decision class to which $R_{i}$ and $R_{k}$ belong and does not have MV. Based on the type of the attribute to which the column $C_{l}$ belongs, the following situations arise:\\
\begin{itemize}
  \item [(i)] Elements of the column $C_{l}$ of $S$ are of categorical (nominal) type:\\
\noindent We now express \emph{A} as a disjoint union of non-empty subsets of \emph{A}, say $B_{\gamma_{p_{1}}},B_{\gamma_{p_{2}}},\ldots,B_{\gamma_{p_{s}}}$ obtained in such a manner that every element of \emph{A} belongs to one of these subsets and no element of \emph{A} is a member of more than one subset of \emph{A}. That is $\emph{A}=B_{\gamma_{p_{1}}}\bigcup B_{\gamma_{p_{2}}}\bigcup, \ldots,\bigcup B_{\gamma_{p_{s}}}$, in which $\gamma_{p_{1}}, \gamma_{p_{2}}, \ldots, \gamma_{p_{s}}$  denote the cardinalities of the  respective subsets $B_{\gamma_{p_{1}}},B_{\gamma_{p_{2}}},\ldots,B_{\gamma_{p_{s}}}$ formed out of the set \emph{A}, with the property that each member of the same subset has the same first co-ordinate and members of no two different subsets have the same first co-ordinate. We define an index
\begin{eqnarray*}
    I_{C_{l}}(R_{i},R_{k})=\left\{
                             \begin{array}{ll}
                              \min\{\frac{\gamma_{p_{i}}}{\gamma_{q_{k}}},\frac{\gamma_{q_{k}}}{\gamma_{p_{i}}}\}, & \hbox{for $i \neq k$;} \\
                               0, & \hbox{otherwise.}
                             \end{array}
                           \right.
\end{eqnarray*}

\noindent where $\gamma_{p_{i}}$ represents the cardinality of the subset $B_{\gamma_{p_{i}}}$, all of whose elements have first co-ordinates $A_{il}$ and $\gamma_{q_{k}}$ represents the cardinality of that subset $B_{\gamma_{q_{k}}}$, all of whose elements have first co-ordinates $A_{kl}$.
 \\
 \item [(ii)] Elements of the column $C_{l}$ of S are of numeric type:\\
\noindent Numeric types can be classified further as integers (whole numbers) or real (fractional numbers). If the attribute is of integer type then we follow the procedure discussed in Case I item (i). For fractional numbers we construct the index $I_{C_{l}}(R_{i},R_{k})$, based on the ratio of the values of the elements  $A_{il},A_{kl}$ of $l^{th}$ column to the mean of the set of elements belonging to \emph{A} that do not have MV and is given by
\begin{eqnarray*}
    I_{C_{l}}(R_{i},R_{k})=\left\{
                             \begin{array}{ll}
                              \min\{\frac{A_{il}}{A^{\#}},\frac{A_{kl}}{A^{\#}}\}, & \hbox{for $i \neq k$;} \\
                               0, & \hbox{otherwise.}
                             \end{array}
                           \right.
\end{eqnarray*}
\noindent In the above definition $A^{\#}$ denotes the average of the $l^{th}$ column entries of all the elements of the set \emph{A} excluding those with MV in the $l^{th}$ column.
\end{itemize}

\item [Case II:]~$A_{in}\neq A_{kn}$ \\
Clearly $R_{i}$ and $R_{k}$ belong to two different decision classes. Consider the subsets $P_{i}$ and $Q_{k}$ consisting of members of \emph{S} that share the same decision with $R_{i}$ and $R_{k}$ respectively and does not have MV. Clearly $P_{i}\bigcap Q_{k} = \emptyset$.  Based on the type of the attribute to which the column $C_{l}$ belongs, the following situations arise:\\
\begin{itemize}
  \item[(i)] Elements of the column $C_{l}$ of S are of nominal or categorical type:\\
 \noindent Following the procedure discussed in Case I item (i) we write \emph{P} and \emph{Q} as a disjoint union of non-empty subsets of  $P_{\beta_{1}},P_{\beta_{2}},\ldots,P_{\beta_{r}}$ and $Q_{\delta_{1}},Q_{\delta_{2}},\ldots,Q_{\delta_{s}}$ respectively in which $\beta_{1}, \beta_{2}, \ldots, \beta_{r}$  and $\delta_{1}, \delta_{2}, \ldots, \delta_{s}$ indicate the cardinalities of the respective subsets. We define the indexing measure between the two records $R_{i}$ and $R_{k}$  as
\begin{eqnarray*}
    I_{C_{l}}(R_{i},R_{k})=\left\{
                             \begin{array}{ll}
                               \max\{\frac{\beta_{r}}{\delta_{s}},\frac{\delta_{s}}{\beta_{r}}\}, & \hbox{for $i \neq k$;} \\
                               0, & \hbox{otherwise.}
                             \end{array}
                           \right.
\end{eqnarray*}
where $\beta_{r}$ represents the cardinality of the subset $P_{\beta_{r}}$ all of whose elements have first co-ordinates $A_{il}$ in set $P$ and $\delta_{s}$ represents the cardinality of that subset $Q_{\delta_{s}}$, all of whose elements have first co-ordinates $A_{kl}$ in set $Q$.
\\
\item[(ii)] Elements of the column $C_{l}$ of S are of numeric type:\\
\noindent  If the type of the attribute is integer we follow the procedure discussed in Case II item (i). For fractional numbers we define  the index $I_{C_{l}}(R_{i},R_{k})$ between the two records $R_{i} and R_{k}$ as
\begin{eqnarray*}
    I_{C_{l}}(R_{i},R_{k})=\left\{
                             \begin{array}{ll}
                               \max\{\frac{A_{il}}{\Lambda},\frac{A_{kl}}{\Lambda}\}, & \hbox{for $i \neq k$;} \\
                               0, & \hbox{otherwise.}
                             \end{array}
                           \right.
\end{eqnarray*}
\noindent In the above definition $\Lambda=\min\{P^{\#} ,Q^{\#}\}$ where $P^{\#},$ and $Q^{\#}$ denote the average of the first column entries of all the elements of the sets $\emph{ P}$ and $\emph{Q}$ excluding those with MV in the $l^{th}$ column.
\end{itemize}
\end{description}

The proximity or distance scores between the clinical records in the data set $S$ can be represented as $D=\{\{0,d_{12},\ldots,d_{1m}\};\{d_{21},0,\ldots,d_{2m}\};\ldots;\{d_{m1},d_{m2},\ldots,0\}\}$ where $d_{ik}=\sqrt{\sum_{l=1}^{n-1}I_{C_{l}}^{2}(R_{i},R_{k})}$. For each of the missing value instances in a  record $R_{i}$ our imputation procedure first computes the score $z(d_{ij})=\frac{(d_{ij}-\overline{d})}{\sqrt{ \frac{1}{m-1} \sum_{i=1}^{m}(d_{ij}-\overline{d})}}$  where $j=1,2\ldots,m$ and $\overline{d}$ denotes the mean distance. We then pick up only those records (nearest neighbors) which satisfy the condition $z(d_{ij}) \leq 0$ where $\{d_{i1},d_{i2},\ldots,d_{im} \}$ denote the distances of the current record $R_{i}$ to all other records in the data set $S$. If the type of attribute is categorical or integer, then the data value that has the highest frequency (mode) of occurrence in the corresponding columns of the nearest records is imputed. For the data values of type real we impute the mean of data values in the corresponding columns of the nearest records.

\textbf{Illustrative example:} The following example illustrates the spirit of the new imputation algorithm. Consider a data set represented by the matrix $S$ consisting of rows $R_{1}$=(?, 12.0, positive), $R_{2}$=( yes, 10.5, positive), $R_{3}$=( no, 14.0, positive) and $R_{4}$=(no, 13.0, negative). The missing value instance ('?') in this data set is present in record $R_{1}$ and column $C_{1}$.  These rows correspond to the data records of four individuals. Clearly the Case I item (i) of the imputation algorithm applies to this data set for determining the missing value. The matrix of the indexing measure $I$ has the following rows: (0,0.86) and (0,0.99) in which $\gamma_{p}=0,$ $\gamma_{q}=1$ and $A^{\#}=12.17$. The relative distances between $R_{1}$ and the other records are computed as $\{0.93,0,0\}$ and the corresponding z-scores are obtained as $\{-0.57,-0.57,1.154\}$.  Since z $\le$ 0 for the distances between $R_{1}$ and $R_{2}$ and also $R_{1}$ and $R_{3}$, we conclude that the records $R_{2}$ and $R_{3}$ are nearer to $R_{1}$ and hence the highest frequency (mode) of the data value in column $C_{1}$ is 'yes'. Accordingly this value is a suitable candidate for imputation.
\subsection{Identification of influential features}
\label{FS}
 In situations presented by real world processes, influential features are often unknown \emph{a priori}, hence features that are redundant or those that are weakly participating in decision making must be identified and appropriately handled. The features selection procedures can be categorized as random or sequential. The sequential methods such as forward selection, backward elimination and bidirectional selection employ greedy methods and hence may not often be successful in finding the optimal features subsets. In contrast to this stochastic optimization methods such as genetic algorithms (GAs) perform global search and are capable of effectively exploring large search spaces~\cite{Goldberg1989}. In our approach we adopt a wrapper subset based feature evaluation model~\cite{Ron1997} where the method of classification itself is used to measure the importance of the features sub set  identified by the GA.
\subsection{Predictive modeling using decision trees}
 An alternating decision tree (ADT) consists of decision nodes (splitter node) and prediction nodes which can either be an interior node or a leaf node. The tree generates a prediction node at the root and then alternates between decision nodes and further prediction nodes. Decision nodes specify a predicate condition and prediction nodes contain a single number denoting the predictive value. An instance can be classified by following all paths for which all decision nodes are true and summing the relevant prediction nodes that are traversed. A positive sum implies membership of one class and the negative sum indicates the membership of the opposite class.
\section{Performance evaluation methods}
\label{evalmethods}
The standard definitions of the performance measures such as the specificity (SP), sensitivity (SE), receiver operator characteristics (ROC) and area under ROC (AUC) based on number of true positives, true negatives, false positives and false negatives are utilized in our experimental analysis.
We employed a stratified $k$-fold cross validation for estimating the test error on classification algorithms. We have randomly divided the given data set into $k$ disjoint subsets. Each subset is roughly of equal size and has the same class proportions as in the original data set. The classification model has been built by setting aside one of the subsets as test data set and train the classifier using the other nine subsets. The trained model is then employed in classifying the test data set. The experiment is repeated by setting aside each of the $k$ subsets as test data sets one at a time.
To compute ROC for $k$ folds we first train a classifier using the training data set of a $k$ fold and then obtain the scores in terms of the predicted probabilities for positives and negatives from the trained classifier using the test data set corresponding to the same fold as the training data. Once all the probabilities and corresponding actual decisions are collected, the ROC is obtained by first computing the thresholds using the quartiles of the cumulative predictive probabilities of all the $k$ folds. For each threshold value the measures SE and SP are computed. The false positive rate and true positive rate values of the ROC is taken as (1-SP) and SE respectively. The AUC is computed by applying a trapezoidal rule on the data points of the ROC curve. The optimal cut off or operating point is the threshold that is closest point to (0,1) on the ROC curve which gives the equal error rate. The optimal values of AUC, SE, SP are computed for this cut off point.
\begin{algorithm}[!t]\scriptsize
\caption{The NM Methodology}
\label{chap9:NM}
\begin{algorithmic}
\Require
\begin{enumerate}[label=(\alph{*})]
            \item Data sets for the purpose of decision making $S(m,n)$ where $m$ and $n$ are number of records and attributes respectively and the members of $S$ may have MV in any of the attributes except in the decision attribute, which is the last attribute in the record.
            \item  The type of attribute $C$ of the columns in the data set.
\end{enumerate}
\Ensure
\begin{enumerate}[label=(\alph{*})]
            \item  Classification accuracy for a given data set $S$.
            \item  Performance metrics AUC, SE, SP.
\end{enumerate}
   \textbf{Algorithm}
   \begin{enumerate}[label=(\arabic{*})]
        \item Identify and collect all records in a data  set $S$
        \item  Impute the MV in the data set $S$ using the procedure discussed in Section~\ref{chap9:RNI}.
        \item Extract the influential features using a wrapper based approach with  genetic search for identifying features subsets and alternating decision tree for its evaluation as discussed in Section~\ref{FS}.
       \item Split the dataset in to training and testing sets using a stratified $k$ fold cross validation procedure. Denote each training and testing data set by $T_{k}$ and $R_{k}$ respectively.
   \item For each $k$ compute the following
      \begin{enumerate}[label=(\roman{*})]\label{NMproc}
        \item Build the ADT using the records obtained from $T_{k}$. \label{buildadt}
        \item Compute the predicted probabilities (scores) for both positive and negative diagnosis of dengue from the ADT built in Step (5)-(i) using the test data set $R_{k}$. Designate the set consisting of all these scores by $P$.
        \item Identify and collect the actual diagnosis from the test data set $R_{k}$ in to set denoted by $L$.\label{step3}
     \end{enumerate}
        \item  Repeat the Steps (5)-(i) to Step (5)-(iii) for each fold.
        \item Obtain the performance metrics AUC, SE and SP utilizing the sets $L$ and $P$.
         \item RETURN AUC, SE, SP.
         \item END.
   \end{enumerate}
\end{algorithmic}
\end{algorithm}
\section{Experiments and results}
\label{chap8:sec3}
In our methodology we have employed a stratified ten-fold cross validation ( $k=10$) procedure. We applied a standard implementation of SVM with radial basis function kernel~\cite{gomes2010} using LibSVM package~\cite{chang2001}. The GA algorithm for features selection has been performed using the parameter values: cross over probability=$1.0$ and mutation probability=$0.001$. The standard implementation of C4.5, LOR algorithms in Weka$^\copyright$~\cite{Witten2005} are considered for evaluating the performance of our algorithm. We have implemented the NM algorithm and the performance evaluation methods in Matlab$^\copyright$. A non-parametric statistical test proposed by Wilcoxon~\cite{Wilcoxon1945} is used to compare the performances of the algorithms. We compared the NM with the state-of-the-art methodologies employed in diagnosis of dengue fever using different performance measures discussed in Section~\ref{evalmethods}.
\subsection{Data sets}
We have obtained four surveillance data sets from case-patients admitted into hospitals located in central and western States of India. Standard procedures were adopted in collecting the clinical and demographic attributes of the patients. The probable cases of the dengue fever are arrived through definitive laboratory tests such as ELISA. The patients records include clinical symptoms: fever, fever duration, headache, retro-orbital pain (eye pain), myalgia (body pain), arthralgia (joint pain), nausea or vomiting, bleeding gums, rash,  bleeding sites, restlessness and  abdominal pain and laboratory features: haemoglobin (Hb), white blood cell count (WBC), packed cell volume (PCV) and platelets. The last attribute in data set is the decision attribute. The clinical records are then re-grouped into four data sets. The first data set (DS1) comprises of $646$ adults (age$\geq$ $16$ years) with clinical symptoms and laboratory features out of which $256$ were dengue positive and $390$ are dengue negative. The second data set (DS2) is a part of DS1 consisting of only clinical symptoms (ignoring the laboratory features) and has the same number of records as in DS1. The third data set (DS3) consists of $398$ children (age between $5-15$ years)~\cite{ramos2009} with clinical symptoms and laboratory features, out of which $93$ were dengue positive and $305$ were dengue negative. The fourth data set (DS4) is a part of DS3 with only clinical symptoms and has same number of records as DS3.
\subsection{Results}
The performance of the NM is compared with other methodologies (C4.5, SVM and LOR) on the data sets used in the present study and the classification accuracies are presented in Table~\ref{chap8:tab1}. A hundred percent accuracy is reported by NM both in data sets DS1 and DS3.
\begin{table}[!t]\scriptsize
\centering
\caption{Performance comparison of the NM with other methodologies (C4.5, SVM and LOR) on the data sets used in the present study}
\label{chap8:tab1}
\begin{tabular}{c||c||c||c||c||c}
\hline
\bf{Dataset}&\bf{Method}&\bf{Accuracy}&\bf{SE}&\bf{SP}&\bf{AUC} \\
&&(\%)&&&\\
\hline \hline
\multirow{6}{*}{DS1}&\bf NM&\bf 100.00&\bf 100.00&\bf 100.00&\bf 1.00\\
&C4.5& 96.44& 95.90& 97.27&  1.00\\
&LOR& 91.02& 89.49& 93.36&  0.96\\
&SVM& 96.75& 97.18& 96.09&  0.97\\
\hline
\multirow{6}{*}{DS2}&\bf NM&\bf 86.53&\bf 88.97&\bf 82.81& \bf 0.93\\
&C4.5& 82.35& 87.18& 75.00&  0.84\\
&LOR& 72.91& 74.36& 70.70&  0.78\\
&SVM& 78.17& 89.49& 60.94&  0.75\\
\hline
\multirow{6}{*}{DS3}&\bf NM&\bf 100.00&\bf 100.00&\bf 100.00&\bf 1.00\\
&C4.5& 94.97& 95.41& 93.55&  0.99\\
&LOR& 92.71& 92.79& 92.47&  0.96\\
&SVM& 98.99& 98.69&100.00&  0.99\\
\hline
\multirow{6}{*}{DS4}&\bf NM&\bf 95.48&\bf 98.03&\bf 87.10&\bf  0.95\\
&C4.5& 90.20& 91.48& 86.02&  0.91\\
&LOR& 88.44& 89.84& 83.87&  0.90\\
&SVM& 92.71& 98.03& 75.27&  0.87\\
\hline
\end{tabular}
\end{table}
The Wilcoxon matched-pairs rank sum test results comparing the accuracies of NM with other methodologies are shown in Table~\ref{chap8:tab2}. For example, the positive rank sum of $55.0$ and negative rank sum of $0.0$ with a p-value$<0.01$ for C4.5 using data set DS1 (first row Table~\ref{chap8:tab2}) indicates the superior performance of the new methodology over C4.5 and also in respect of other methods as well.
\begin{table}[H]\scriptsize
\centering
\caption{Wilcoxon matched-pairs rank sum test for comparing the performance of NM with other methodologies used in diagnosis of dengue fever} \label{chap8:tab2}
\begin{tabular}{c||c||c||c}
\hline
\bfseries Dataset&\bfseries Method&\bfseries Rank sum(+, -)&\bfseries p-value\\ \hline \hline
\multirow{5}{*}{DS1} & C4.5&55.0, 0.0&0.002\\
&LOR&55.0, 0.0&0.002\\
&SVM&45.0, 0.0&0.004\\
\hline
\multirow{5}{*}{DS2} & C4.5&55.0, 0.0&0.002\\
&LOR&55.0, 0.0&0.002\\
&SVM&55.0, 0.0&0.002\\
 \hline
\multirow{5}{*}{DS3} & C4.5&36.0, 0.0&0.008\\
&LOR&36.0, 0.0&0.008\\
&SVM&10.0, 0.0&0.125\\
 \hline
\multirow{5}{*}{DS4} & C4.5&38.5, 6.5&0.074\\
&LOR&37.0, 8.0&0.098\\
&SVM&27.0, 9.0&0.25\\
 \hline
\end{tabular}
\end{table}
The above comparisons and statistical tests clearly demonstrate the significance of our methodology in identifying the suspected dengue both in children and adults. The imputation strategy employed in our methodology has improved the classification accuracies  when compared with C4.5 which uses a modified information gain measure to generate the decision tree in presence of MV. The mean imputation strategies adopted in SVM and LOR could not render classification accuracies higher than NM.

\begin{figure*}[!t]
\hspace*{1\baselineskip}
\centerline{\subfloat[DS1]{\includegraphics[width=0.32\textwidth]{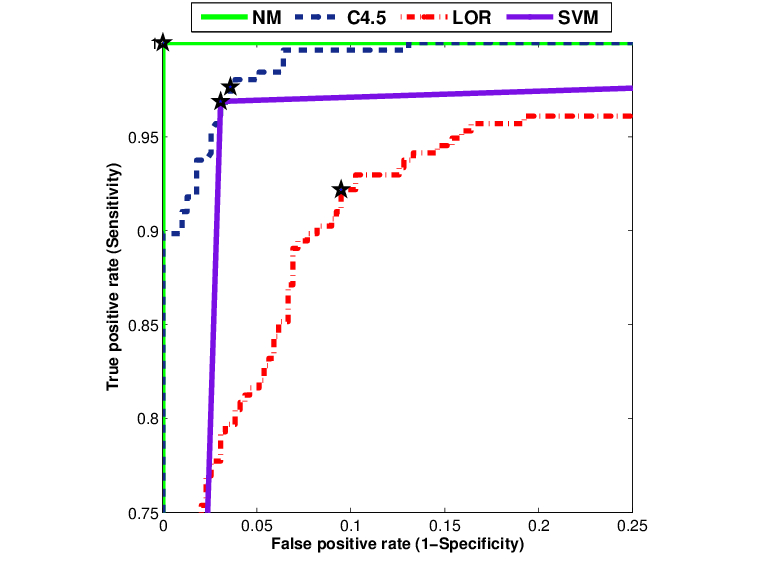}
\label{chap8:fig2}}
\hspace*{-2\baselineskip}
\subfloat[DS2]{\includegraphics[width=0.32\textwidth]{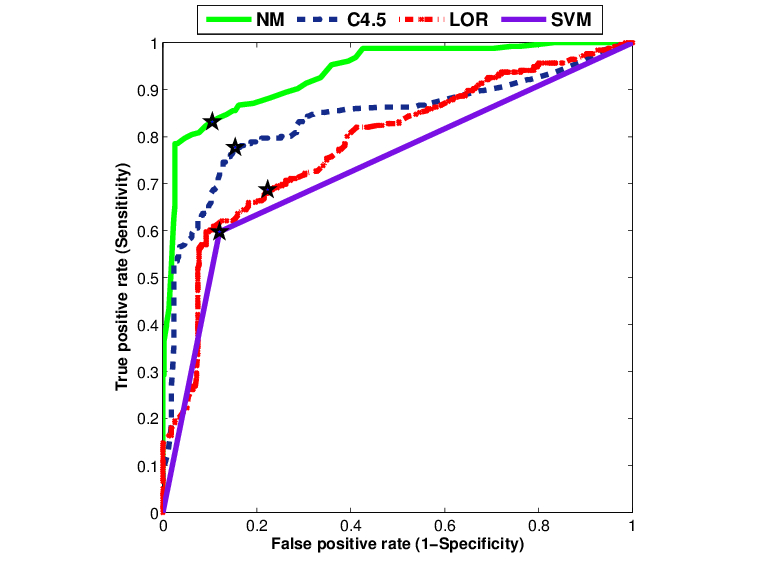}
\label{chap8:fig3}}
\hspace*{-2\baselineskip}
\subfloat[DS3]{\includegraphics[width=0.32\textwidth]{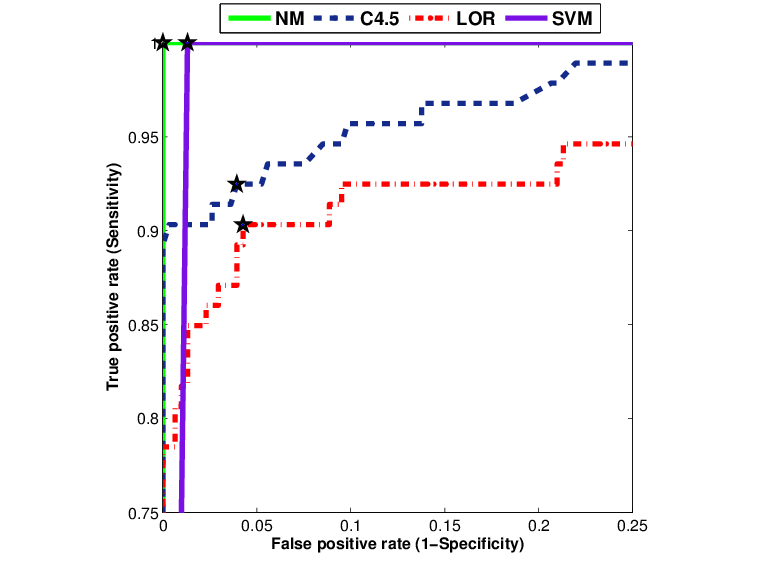}
\label{chap8:fig4}}
\hspace*{-2\baselineskip}
\subfloat[DS4]{\includegraphics[width=0.32\textwidth]{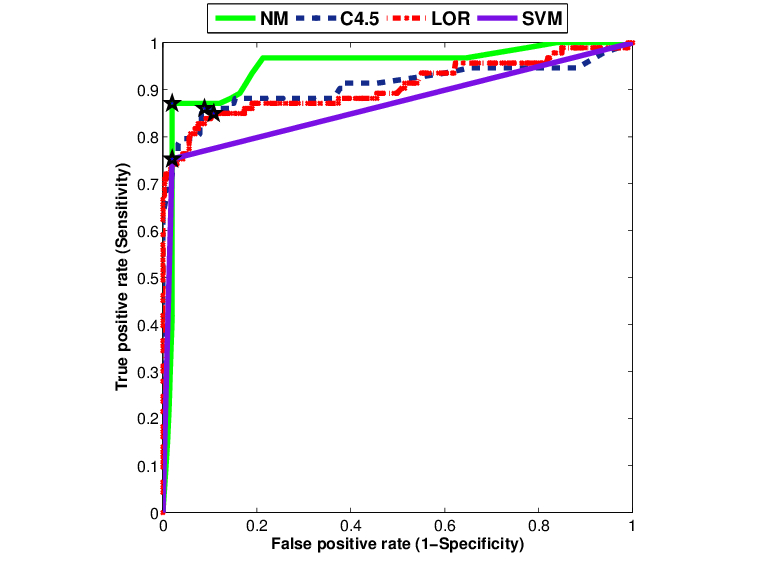}
\label{chap8:fig5}}}
\caption{ROC curves}
\label{fig_sim}
\end{figure*}

The features subsets identified by the NM is shown in Table~\ref{chap8:tab3}. The application of features selection method reduced the number of attributes by $75\%$ in DS1 and $87.5\%$ in DS3 data sets. Our methodology identified some of the clinical symptoms and laboratory features in adults (vomiting and abdominal pain) different from those in children which are in concurrence with earlier studies~\cite{Enid2003,Ole2004}. The clinical attribute rash was identified as an important feature in adults but not in children. This may be explained by the relative frequency of the secondary infections in adults~\cite{Cobra1995}. Arthralgia was found to be influencing the final diagnosis of dengue both in children and adults.
\begin{table}[!t]\scriptsize
\centering
\caption{Influential features subsets identified by NM} \label{chap8:tab3}
\begin{tabular}{p{0.2in}||c||p{0.5in}||c||p{1.2in}}
\hline
\bf{Data set} & \bf{\# Orignal}&\bf{\# influential features}&\bf{Accuracy}&\bf{features} \\
&\bf{features}&&\bf{(\%)}& \bf{identified}\\
\hline \hline
DS1&16&5&100.00&retro-orbital pain , arthralgia, fever duration, platelet, fever\\ \hline
DS2&9&6&86.53&vomiting or nausea, myalgia, rash, bleeding sites, abdominal pain, arthralgia\\ \hline
DS3&16&2&100.00&Hb, fever\\ \hline
DS4&9&2&95.48&retro-orbital pain, arthralgia\\ \hline
\end{tabular}
\end{table}
The ROC curves comparing the performance of NM with other methodologies are shown in Figs.~\ref{chap8:fig2}-\ref{chap8:fig5}. The operating point or cut off point ($p<0.001$) is shown as a pentagon on each of the ROC curves. The ROC curves clearly demonstrate the superior performance of NM over other methods used in the diagnosis of dengue.
\section{Performance comparison of new imputation algorithm with bench marking data sets}
\label{perMV}
Since no specific studies on imputation of missing values in dengue data sets we have utilized some bench marking data sets obtained from Keel and University of California Irvin (UCI) machine learning data repositories~\cite{Alcala-Fdez2010,Frank2010} to test the performance of the new imputation algorithm. The Wilcoxon statistics in Table~\ref{chap8:tab5} is computed based on the accuracies obtained by the new imputation algorithm with the accuracies of those obtained by other imputation algorithms using a C4.5 decision tree. The results in Table~\ref{chap8:tab5} clearly demonstrate the fact that our algorithm is superior to other imputation algorithms as the positive rank sums are higher than the negative rank sums ($p<0.05$) in all the cases.
\begin{table}[H]\scriptsize
\caption{Wilcoxon sign rank statistics for matched pairs comparing the new imputation algorithm with other imputation methods using C4.5 decision tree}
\label {chap8:tab5}
\centering
\begin{tabular}{l||c||c||c||c}
\hline
\bfseries Method	&	\bfseries Rank Sums &\bfseries Test& \bfseries Critical& \bfseries p-value \\
& \bfseries (+, -)&\bfseries  Statistics&\bfseries Value&\\
\hline \hline
FKMI	&	78.5, 12.5&	12.5&	18	&0.021\\ \hline
KMI		&85.0, 6.0	&6	&18	&0.003\\ \hline
KNNI	&	76.0, 15.0&	15&	18	&0.032\\ \hline
WKNNI	&	83.0, 8.0	&8	&18	&0.006\\ \hline
\end{tabular}
\end{table}

\section{Computational complexity}
\label{chap8:sec4}
\begin{figure}[!t]
\centering
\includegraphics[width=0.35\textwidth]{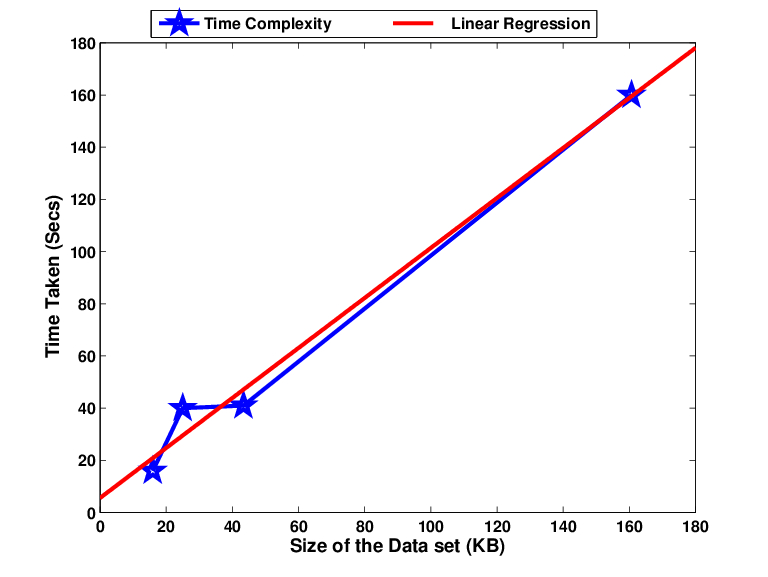}\\
  \caption{Computational complexity of the NM}\label{chap8:fig6}
\end{figure}

The computational complexity is a measure of the performance of the algorithm. For each data set having $n$ attributes and $m$ records, we select only those subset of records $m_{1} \le m$, in which missing values are present. The distances  are computed for all attributes $n$ excluding the decision attribute. So, the time complexity for computing the distance would be $O(m_{1}*(n-1))$. The time complexity for selecting the nearest records is of order $O(m_{1})$.  For computing the frequency of occurrences for nominal attributes and average for numeric attributes the time taken would be of the order $O(m_{1})$. Therefore, for a given data set with $k$-fold cross validation having $n$ attributes and $m$ records, the time complexity of our new imputation algorithm would be $k*(O(m_{1}*(n-1)*m)+2*O(m_{1}))$ which is asymptotically linear. Our experiments were conducted on a personal computer having a Intel(R) core (TM) 2 Duo,  CPU @$2.93$ GHZ  processor with $4$ GB RAM.  For each data set the computational time for imputation and features selection is measured in terms of the number of CPU clock cycles elapsed in seconds. Based on the results, we obtain a scatter plot (red line in Fig.~\ref{chap8:fig6}) between the varying database sizes and the time taken by NM. Also, we employed a linear regression on our results and obtained the relation between the time taken (T) and the data size (D) as $T=0.96D+5.54$, $\alpha=0.05$, $p<0.05$, $r^2=0.98$. The presence of the linear trend between the time taken and the varying database sizes ensures the numerical scalability of the performance of NM in terms of asymptotic linearity.
\section{Comparison of related methodologies on dengue studies}
\label{chap8:sec5}
In this section we compare the results (Table~\ref{chap8:tab4}) obtained in~\cite{Tanner2008,ramos2009,Chadwick2006} with the results of our new methodology on the current data set of $1044$ individuals including children and adults. As compared to~\cite{ramos2009} where children with rash were having SE of $41.2\%$ and SP of $95.5\%$ our methodology when applied on the data set DS2 resulted in an accuracy of $86.53\%$, SE of $88.97\%$ and SP of $82.81\%$ which is considered to be a good classification model as both SE and SP are higher than $80\%$. In~\cite{Tanner2008} both clinical and laboratory features were utilized to develop decision rules using C4.5 decision tree and they have reported a SE of $87.8\%$ and SP of $75.7\%$. In comparison to~\cite{Tanner2008} our methodology when applied on DS1 and DS3 had resulted in SE of $100\%$ and SP of $100\%$.
\begin{table}[!t]\scriptsize
\centering
\caption{Evaluation of NM with other related methodologies on dengue studies} \label{chap8:tab4}
\begin{tabular}{p{0.6in}||p{0.3in}||p{0.3in}||p{0.3in}||p{0.3in}||p{0.2in}||p{0.2in}}
\hline
\bfseries{State-of-the-art}&\bfseries{\#Patients (DF)}&\bfseries{Records with MV}&\bfseries{Methods}&\bfseries{Accuracy (\%)}&\bfseries{SE (\%)}&\bfseries{SP (\%)}\\
\hline \hline
Chadwick et al.,~\cite{Chadwick2006} (clinical)&381 (148)&deleted&LOR, OR&$84.5$&$84$&$85$ \\ \hline
Chadwick et al.,~\cite{Chadwick2006} (laboratory)&381 (148)&deleted&-do-&$76.5$&$74$&$79$ \\ \hline
Ramos et al.,~\cite{ramos2009} (clinical, children)&938 (38)&manual update&-do-&$68.95$&$41.2$&$95.5$ \\ \hline
Tanner et al.,~\cite{Tanner2008} (laboratory)&1200 (173)&deleted&C4.5&$81.75$&$87.8$&$75.7$\\ \hline
Gomes et al.,~\cite{gomes2010} (gene database)&20 (15)&-&SVM&$85$&-&- \\ \hline
NM (DS1) (adults, clinical \& laboratory) &1044 (256)&imputed (new algorithm)&ADT, GA &$100$&$100$&$100$ \\ \hline
NM (DS2) (adults, clinical) &1044 (256)&-do-&-do- &$86.53$&$88.97$&$82.81$ \\ \hline
NM (DS3) (children, clinical \& laboratory) &1044 (93)&-do-&-do-&$100$&$100$&$100$ \\ \hline
NM (DS4) (children, clinical) &1044 (305)&-do-&-do-&$95.48$&$98.03$&$87.10$ \\ \hline
\end{tabular}
\end{table}
From these comparisons we conclude that the new methodology presented in this study if applied on the data sets used in~\cite{Tanner2008,ramos2009,Chadwick2006} would yield more accurate results.
\section{Conclusions and discussion}
\label{chap8:sec6}
A new methodology (NM) with built in features for imputation of missing values and identification of influential attributes is discussed. The NM has out performed the state-of-the-art methodologies in diagnosis of dengue fever on all the four data sets considered in our experiments. The NM has generated a decision tree with an accuracy of $100.0\%$ in children and adults using both clinical and laboratory features. Based on the performance measures we conclude that the use of the new imputation strategy and features selection methods with wrapper based subset evaluation using genetic search has improved the accuracies of the predictions. Though the new methodology discussed in this paper may be taken as a universal tool for the effective diagnosis of this disease, it remains to be seen whether or not this methodology is geographically independent. However, we are willing to share our predictive methodologies and strategies with the researchers working on dengue fever all over the globe. We hold the view that more intensive and introspective studies of this kind will pave way for better clinical management and virological surveillance of dengue fever.
\section*{Acknowledgments}
We thank the Associate Editor and the anonymous reviewers for their constructive suggestions on our paper. This research is supported by the Foundation for Scientific Research and Technological Innovation (FSRTI)- A Constituent Division of Sri Vadrevu Seshagiri Rao Memorial Charitable Trust, Hyderabad - 500 035, India.

\bibliographystyle{IEEETran}

\vspace*{-2\baselineskip}
\end{document}